\documentclass[11pt,a4paper]{article}
\usepackage[hyperref]{naaclhlt2018}
\usepackage{times}
\usepackage{latexsym}
\usepackage{graphicx}
\usepackage{url}
\usepackage{multirow}
\usepackage{amsmath}
\usepackage{amsfonts,amssymb}
\usepackage{subfigure}
\aclfinalcopy 

\title{SJTU-NLP at SemEval-2018 Task 9: \\ Neural Hypernym Discovery with Term Embeddings}

\author{Zhuosheng Zhang$^{1,2}$, Jiangtong Li$^{3}$, Hai Zhao$^{1,2,}$\thanks{$\ $ Corresponding author. This paper was partially supported by
		National Key Research and Development Program of China (No. 2017YFB0304100),
		National Natural Science Foundation of China (No. 61672343 and No. 61733011),
		Key Project of National Society Science Foundation of China (No. 15-ZDA041),
		The Art and Science Interdisciplinary Funds of Shanghai Jiao Tong University (No. 14JCRZ04).} , Bingjie Tang$^{4}$  \\
	$^{1}$Department of Computer Science and Engineering, Shanghai Jiao Tong University \\
	$^{2}$Key Laboratory of Shanghai Education Commission for Intelligent Interaction \\ and Cognitive Engineering, Shanghai Jiao Tong University, Shanghai, 200240, China\\
	$^{3}$College of Zhiyuan, Shanghai Jiao Tong University, China\\
	$^{4}$School of Computer, Huazhong University of Science and Technology, China\\
	{\tt \{zhangzs, keep\_moving-lee\}@sjtu.edu.cn, zhaohai@cs.sjtu.edu.cn, } \\ {\tt alexistang@foxmail.com}
}

\begin{document}
	\maketitle
	\begin{abstract}
	This paper describes a hypernym discovery system for our participation in the SemEval-2018 Task 9, which aims to discover the best (set of) candidate hypernyms for input concepts or entities, given the search space of a pre-defined vocabulary. We introduce a neural network architecture for the concerned task and empirically study various neural network models to build the representations in latent space for words and phrases. The evaluated models include convolutional neural network, long-short term memory network, gated recurrent unit and recurrent convolutional neural network. We also explore different embedding methods, including word embedding and sense embedding for better performance.  
	\end{abstract}

	\section{Introduction}
	
	Hypernym-hyponym relationship is an \emph{is-a} semantic relation between terms as shown in Table \ref{tab:example}. Various natural language processing (NLP) tasks, especially those semantically intensive ones aiming for inference and reasoning with generalization capability, such as question answering \cite{Harabagiu2006Methods, Yahya2013Robust} and textual entailment \cite{Dagan2013Recognizing,Roller2016Relations}, can benefit from identifying semantic relations between words beyond synonymy.
	
	The \emph{hypernym discovery} task \cite{semeval2018task9} aims to discover the most appropriate hypernym(s) for input concepts or entities from a pre-defined corpus. A relevant well-known scenario is \emph{hypernym detection}, which is a binary task to decide whether a hypernymic relationship holds between a pair of words or not. A hypernym detection system should be capable of learning taxonomy and lexical semantics, including pattern-based methods \cite{Boella2013Supervised,Espinosa2016ExTaSem} and graph-based approaches \cite{Fountain2012Taxonomy,Velardi2013OntoLearn,Kang2016TaxoFinder}. However, our concerned task, hypernym discovery, is rather more challenging since it requires the systems to explore the semantic connection with all the exhausted candidates in the latent space and rank a candidate set instead of a binary classification in previous work. The other challenge is representation for terms, including words and phrases, where the phrase embedding could not be obtained by word embeddings directly. A simple method is to average the inner word embeddings to form the phrase embedding. However, this is too coarse since each word might share different weights. Current systems like \cite{Espinosa2016Supervised} commonly discover hypernymic relations by exploiting linear transformation matrix in embedding space, where the embedding should contain words and phrases, resulting to be parameter-exploded and hard to train. Besides, these systems might be insufficient to obtain the deep relationships between terms. 
	
	\begin{table}[h]\small
	\centering
	{\small
		\begin{tabular}{|l|p{5.4cm}|}
		\hline
		Hyponym & Hypernyms \\
		\hline
		Heming & actor, person, company\\
		\hline
		Kralendijk & town, city, provincial capital, capital\\
		\hline
		StarCraft & video game, pc game, computer game, videogaming, comic,	electronic game, scientifiction \\
		\hline
		\end{tabular}
	}
	\caption{\label{tab:example} Examples of hypernym-hyponym relationship.}
	\end{table}
	
	Recently, neural network (NN) models have shown competitive or even better results than traditional linear models with handcrafted sparse features \cite{Qin2016A,Pang2016I,Qin2016Shallow,Wang2016Connecting,Zhao2017A,Wang2017A,Qin2017Adversarial,Cai2017Pair,Zhao2017B,Li2018}. In this work, we introduce a neural network architecture for the concerned task and empirically study various neural networks to model the distributed representations for words and phrases. 
	
	In our system, we leverage an unambiguous vector representation via term embedding, and we take advantage of deep neural networks to discover the hypernym relationships between terms.
	
	The rest of the paper is organized as follows: Section 2 briefly describes our system, Section 3 shows our experiments on the hyperym discovery task including the general-purpose and domain-specific one. Section 4 concludes this paper.
	
	\section{System Overview}
	Our hypernym discovery system can be roughly split into two parts, \emph{Term Embedding} and \emph{Hypernym Relationship Learning}. We first train term embeddings, either using word embedding or sense embedding to represent each word. Then, neural networks are used to discover and rank the hypernym candidates for given terms.
	\subsection{Embedding}
	To use deep neural networks, symbolic data needs to be transformed into distributed representations\cite{Wang2016Learning,Qin2016A,Cai2016Neural,Zhang2016Probabilistic,Wang2016Converting,Wang2015Bilingual,Cai2017Fast}. We use \emph{Glove} toolkit to train the word embeddings using \emph{UMBC} corpus \cite{Han2013UMBC}. Moreover, in order to perform word sense induction and disambiguation, the word embedding could be transformed to sense embedding, which is induced from exhisting word embeddings via clustering of ego-networks \cite{pelevina-EtAl:2016:RepL4NLP} of related words. Thus, each input word or phrase is embedded into vector sequence, $w = \{x_{1},x_{2},\dots,x_{l}\}$ where $l$ denotes the sequence length. If the input term is a word, then $l=1$ while for phrases, $l$ means the number of words.
	
	\subsection{Hypernym Learning}
	Previous work like TAXOEMBED \cite{Espinosa2016Supervised} uses transformation matrix for hypernm relationship learning, which might be not optimal due to the lack of deeper nonlinear feature extraction. Thus, we empirically survey various neural networks to represent terms in latent space. After obtaining the representation for input term and all the candidate hypernyms, to give the ranked hypernym list, the cosine similarity between the term and the candidate hypernym is computed by,
	\begin{align*}
	cosine = \frac{\sum_{i=1}^{n}(x_{i}\times y_{i})}{\sum_{i=1}^{n}x_{i}^{2} \times  \sum_{i=1}^{n}y_{i}^{2}}
	\end{align*} 
	where $x_{i}$ and $y_{i}$ denote the two concerned vectors.
	Our candidate neural networks include Convolutional Neural Network (CNN), Long-short Term Memory network (LSTM), Gated Recurrent Unit (GRU) and Recurrent Convolutional Neural Network (RCNN).
	
	\paragraph{GRU} The structure of GRU \cite{Cho2014Learning} used in this paper are described as follows.
	\begin{align*}
		r_{t} &=\sigma (W_{r}x_{t}+U_{r}h_{t-1}+b_{r}), \\
		z_{t} &=\sigma (W_{z}x_{t}+U_{z}h_{t-1}+b_{z}), \\
		\tilde{h}_{t} &=\textup{tanh}(W_{h}x_{t}+U_{h}(r{t}\odot h_{t-1})+b_{h}) \\
		h_{t} & = (1-z_{t})\odot h_{t-1}+z_{t}\odot \tilde{h}_{t}
	\end{align*}
	where $\odot$ denotes the element-wise multiplication. $r_{t}$ and $z_{t}$ are the reset and update gates respectively, and $\tilde{h}_{t}$ the hidden states. 
	
	\paragraph{LSTM}
	LSTM \cite{Hochreiter1997Long} unit is defined as follows.
	\begin{align*}
	&i_{t}=\sigma (W_{i}{x}_{t}+{W}_{h}{h}_{t-1}+{b}_{i}),\\
	&{f_{t}}=\sigma ({W}_{f}{x}_{t}+{W}_{f}{h}_{t-1}+{b}_{f}),\\
	&{u_{t}}=\sigma ({W}_{u}{x}_{t}+{W}_{u}{h}_{t-1}+{b}_{u}),\\
	&{c_{t}}={f}_{t}\odot {c}_{t-1}+{i}_{t}\odot\tanh({W}_{c}{x}_{t}+{W}_{c}{h}_{t-1}+{b}_{c}),\\
	&{h_{t}}=\tanh ({c}_{t})\odot {u}_{t},
	\end{align*}
	where $\sigma$ stands for the sigmoid function, $\odot$ represents element-wise multiplication and $W_{i}, W_{f}, W_{u}, {W}_{c}, {b}_{i}, {b}_{f}, {b}_{u}, {b}_{c}$ are model parameters. ${i_{t}},{f_{t}},{u_{t}},{c_{t}},{h_{t}}$ are the input gates, forget gates, memory cells, output gates and the current state, 
	respectively. 
	
	\paragraph{CNN}
	Convolutional neural networks have also been successfully applied to various NLP tasks, in which the temporal convolution operation and associated filters map local chunks (windows) of
	the input into a feature representation. 
	
	Concretely, let $n$ denote the filter width, filter matrices [$W_1$, $W_2$, \dots , $W_k$] with several variable sizes  [$l_1, l_2,\dots , l_k$] are utilized to perform the  convolution operations for input embeddings. For the sake of simplicity, we will explain the procedure for only one embedding sequence. The embedding will be transformed to sequences $c_j (j \in [1, k])$ : 
	\begin{align*}
	{c}_j  = [\dots; \tanh ({W}_j\cdot {x}_{[i:i+l_j-1]} + {b}_j); \dots]
	\end{align*}
	where $[i:i+l_j-1]$ indexes the convolution window. Additionally, we apply wide convolution operation between embedding layer and filter matrices, because it ensures that all weights in the filters reach the entire sentence, including the words at the margins.
	
	A \emph{one-max-pooling} operation is adopted after convolution and the output vector ${s}$ is obtained through concatenating all the mappings for those $k$ filters.
	\begin{align*}
	&s_{j} = {max}(c_{j}) \\
	&s = [s_{1}\oplus \cdots  \oplus s_{j} \oplus \cdots \oplus s_{k}]
	\end{align*}
	In this way, the model can capture the critical features in the sentence with different filters.
	
	\begin{table*}[t]\centering \small
		{
			\begin{tabular}{l|l|l|l|l|l|l|l}
				\hline
				\hline
				Embedding &	Model & MAP  &  MRR  & P@1 & P@3 & P@5 & P 15 \\
				\hline
				\multirow{3}{*}{Word}  
				& TEA  &  6.10 &  11.13 &  4.00 &  6.00 &  5.40 &  5.14  \\
				& GRU  &  8.13 &  16.22 &  8.00 &  \textbf{8.00} &  6.67 &  6.94  \\
				& LSTM &  3.95 &  7.52 &  4.00 &  4.33 &  3.97 &  3.97  \\
				& CNN  &  7.32 &  13.33 &  \textbf{8.00} &  9.00 &  7.80 &  6.94  \\
				& RCNN &  \textbf{8.74} &  \textbf{12.83} &  6.00 &  \textbf{9.67} &  \textbf{8.87} &  \textbf{9.15}  \\
				\hline
				\multirow{3}{*}{Sense} 
				& TEA &  4.42 &  8.71 &  0.00 &  4.04 &  4.19 &  5.31  \\
				& GRU  &  5.42 &  9.44 &  0.00 &  4.44 &  4.89 &  5.83  \\
				& LSTM &  5.62 &  9.97 &  4.00 &  4.35 &  5.01 &  6.83  \\
				& CNN  &  6.41 &  10.92 &  2.00 &  5.01 &  5.67 &  6.29  \\
				& RCNN &  5.79 &  9.24 &  0.00 &  4.71 &  5.29 &  6.43  \\
				\hline
				\hline
			\end{tabular}
			
		}
		\caption{\label{tab:result_sense_word} Gold standard
			evaluation on general-purpose subtask.}
	\end{table*}
	
\paragraph{RCNN}
Since some input terms are phrases, whose member words share different weights. In RCNN, an adaptive gated decay mechanism is used to weight the words in the convolution layer. Following \cite{Lei:2016}, we introduce neural gates similar $\lambda$ to LSTMs
to specify when and how to average the observed signals. The resulting architecture integrates recurrent  networks with  non-consecutive  convolutions:
\begin{align*}
\lambda & = \sigma (W^{\lambda}x_{t}+U^{\lambda}h_{t-1}+b^{\lambda}) \\
c_{t}^{1} & =\lambda_{t} \odot c_{t-1}^{1} + (1-\lambda_{t})\odot W_{1}x_{t} \\
c_{t}^{2} & =\lambda_{t} \odot c_{t-1}^{2} + (1-\lambda_{t})\odot (c_{t-1}^{1} + W_{2}x_{t}) \\
& \cdots \\ 
c_{t}^{n} & =\lambda_{t} \odot c_{t-1}^{n} + (1-\lambda_{t})\odot (c_{n-1}^{1} + W_{n}x_{t}) \\
h_{t} & = \tanh(c_{t}^{n}+b)
\end{align*}
where $c_t^{1}, c_t^{2}, \cdots, c_t^{n}$ are accumulator vectors that store weighted averages of 1-gram to $n$-gram features.

For discriminative training, we use a max-margin framework for learning (or fine-tuning) parameters $\theta$. Specifically, a scoring function $\varphi(\cdot,\cdot;\theta)$ is defined to measure the semantic similarity between the corresponding representations of input term and hypernym. Let $p=\{p_1,...p_n\}$ denote the hypernym corpus and $h \in p$ is the ground-truth hypernym to the term $t_{i}$, the optimal parameters $\theta$ are learned by minimizing the max-margin loss:  
\begin{align*}
\max \{\varphi(t_{i},p_{i};\theta)-\varphi(t_{i},a;\theta)+\delta(p_{i},a)\}
\end{align*}
where $\delta(.,.)$ denotes a non-negative margin and $\delta(p_{i},a)$ is a small constant when $a\neq p_{i}$ and 0 otherwise. 

\section{Experiment}
In the following experiments, besides the neural networks, we also simply average the embeddings of an input phrase as our baseline to discover the relationship of terms and their corresponding hypernyms for comparison (denoted as \emph{term embedding averaging, TEA}).

\begin{table*}[t]\centering
\small 
{
	\begin{tabular}{l|l|l|l|l|l|l|l|l|l|l|l|l|l}
		\hline
		\hline
		\multirow{2}{*}{Embed} & \multirow{2}{*}{Model}   & \multicolumn{6}{c|}{medical} & \multicolumn{6}{c}{music} \\
		\cline{3-14}
		& &MAP  &  MRR  & P@1 & P@3 & P@5 & P 15  & MAP  &  MRR  & P@1 & P@3 & P@5 & P 15\\
		\hline
		\multirow{4}{*}{Word} 
		& TEA  &  8.91  &  16.77 &  0.00 &  8.79  &  9.41  &  9.39   &
		7.11  &  14.32 &  0.00 &  10.01  & 10.77 &  9.21  \\
		& GRU  &  13.27 &  21.89 &  0.00 &  13.33 &  14.89 &  14.06  &  15.20 &  20.33 &  0.00 &  17.78 &  18.67 &  15.45  \\
		& LSTM &  11.49 &  21.11 &  0.00 &  \textbf{17.78} &  12.22 &  11.83  &  14.08 &  20.77 &  0.07 &  13.33 &  16.00 &  15.00  \\
		& CNN  &  \textbf{18.31} &  \textbf{24.52} &  0.00 &  15.56 &  \textbf{20.44} &  \textbf{20.00}&  \textbf{17.58} &  \textbf{27.15} &  0.00 &  \textbf{20.00} &  \textbf{20.00} &  \textbf{16.04}  \\
		& RCNN &  16.78 &  23.40 &  0.00 &  13.33 &  13.00 &  14.50  &  13.60 &  21.67 &  0.07 &  13.33 &  14.67 &  13.08  \\
		\hline
		\multirow{4}{*}{Sense}
		& TEA  &  2.01 &  4.77 &  0.00 &  2.91 &  2.77 &  3.21  & 2.59 &
		5.28  & 0.00 &  2.12  & 3.01  &  2.93 \\
		& GRU  &  4.88 &  9.11 &  0.00 &  6.67 &  6.42 &  6.91  & 5.32 &  10.74 &  \textbf{2.00} &  4.44 &  5.33 &  4.95  \\
		& LSTM &  5.10 &  10.22 &  0.00 &  6.67 &  6.12 &  6.94  & 4.39 &  10.21 &  0.00 &  8.89 &  5.33 &  3.61  \\
		& CNN  &  4.15 &  7.84 &  0.00 &  4.44 &  6.09 &  6.42  & 4.75 &  9.61 &  0.00 &  6.67 &  6.67 &  4.43 \\
		& RCNN &  4.63 &  9.84 &  0.00 &  6.67 &  6.89 &  6.43  & 4.73 &  8.56 &  0.00 &  4.44 &  6.22 &  4.94  \\
		\hline
		\hline
	\end{tabular}
	
}
\caption{\label{tab:result_domain} Gold standard
	evaluation on domain-specific subtask. ``Embed" is short for ``Embedding".}
\end{table*}

\subsection{Setting}
Our hypernym discovery experiments include general-purpose substask for English and domain-specific ones for medical and music. Our evaluation is based on the following information retrieval metrics: Mean Average Precision (MAP), Mean Reciprocal Rank (MRR), Precision at 1 (P@1), Precision at 3 (P@3), Precision at 5 (P@5), Precision at 15 (P@15).

For the sake of computational efficiency, we simply average the sense embedding if a word has more than one sense embedding (among various domains). Our model was implemented using the Theano\footnote{\url{https://github.com/Theano/Theano}} . The diagonal variant of AdaGrad \cite{Duchi2010Adaptive} is used for neural network training. We tune the hyper-parameters with the following range of values: learning rate $\in \left \{ 1e-3,1e-2 \right \}$, dropout probability $\in \left \{ 0.1, 0.2 \right \}$, CNN filter width $\in \left \{ 2, 3, 4 \right \}$. The hidden dimension of all neural models are 200. The batch size is set to 20 and the word embedding and sense embedding sizes are set to 300. All of our models are trained on a single GPU (NVIDIA GTX 980Ti), with roughly 1.5h for general-purpose subtask for English and 0.5h domain-specific domain-specific ones for medical and music. We run all our models up to 50 epoch and select the best result in validation.

\subsection{Result and analysis}
Table \ref{tab:result_sense_word} shows the result on general-domain subtask for English. All the neural models outperform term embedding averaging in terms of all the metrics. This result indicates simply averaging the embedding of words in a phrase is not an appropriate solution to represent a phrase. Convolution or recurrent gated mechanisms in either CNN-based (CNN, RCNN) or RNN (GRU, LSTM) based neural networks could essentially be helpful of modeling the semantic connections between words in a phrase, and guide the networks to discover the hypernym relationships. We also observe CNN-based network performance is better than RNN-based, which indicates local features between words could be more important than long-term dependency in this task where the term length is up to trigrams.

To investigate the performance of neural models on specific domains, we conduct experiments on medical and medicine subtask. Table \ref{tab:result_domain} shows the result. All the neural models outperform \emph{term embedding averaging} in terms of all the metrics and CNN-based network also performs better than RNN-based ones in most of the metrics using word embedding, which verifies our hypothesis in the general-purpose task. Compared with word embedding, the sense embedding shows a much poorer result though they work closely in general-purpose subtask. The reason might be the simple averaging of sense embedding of various domains for a word, which may introduce too much noise and bias the overall sense representation. This also discloses that modeling the sense embedding of specific domains could be quite important for further improvement.

\section{Conclusion}

In this paper, we introduce a neural network architecture for the hypernym discovery task and empirically study various neural network models to model the representations in latent space for words and phrases. Experiments on three subtasks show the neural models can yield satisfying results. We also evaluate the performance of word embedding and sense embedding, showing that in domain-specific tasks, sense embedding could be much more volatile. 

\bibliography{semeval2018}

\begin{thebibliography}{32}
\expandafter\ifx\csname natexlab\endcsname\relax\def\natexlab#1{#1}\fi

\bibitem[{Boella and Caro(2013)}]{Boella2013Supervised}
Guido Boella and Luigi~Di Caro. 2013.
\newblock Supervised learning of syntactic contexts for uncovering definitions
  and extracting hypernym relations in text databases.
\newblock \emph{Joint European Conference on Machine Learning and Knowledge
  Discovery in Databases}, pages 64--79.

\bibitem[{Cai and Zhao(2016)}]{Cai2016Neural}
Deng Cai and Hai Zhao. 2016.
\newblock Neural word segmentation learning for {Chinese}.
\newblock In \emph{Proceedings of the 54th Annual Meeting of the Association
  for Computational Linguistics (ACL 2016)}, pages 409--420.

\bibitem[{Cai and Zhao(2017)}]{Cai2017Pair}
Deng Cai and Hai Zhao. 2017.
\newblock \emph{Pair-Aware Neural Sentence Modeling for Implicit Discourse
  Relation Classification}.
\newblock IEA/AIE 2017, Part II, LNAI 10351.

\bibitem[{Cai et~al.(2017)Cai, Zhao, Zhang, Xin, Wu, and Huang}]{Cai2017Fast}
Deng Cai, Hai Zhao, Zhisong Zhang, Yuan Xin, Yongjian Wu, and Feiyue Huang.
  2017.
\newblock Fast and accurate neural word segmentation for {Chinese}.
\newblock In \emph{Proceedings of the 55th Annual Meeting of the Association
  for Computational Linguistics (ACL 2017)}, pages 608--615.

\bibitem[{Camacho-Collados et~al.(2018)Camacho-Collados, Delli~Bovi,
  Espinosa-Anke, Oramas, Pasini, Santus, Shwartz, Navigli, and
  Saggion}]{semeval2018task9}
Jose Camacho-Collados, Claudio Delli~Bovi, Luis Espinosa-Anke, Sergio Oramas,
  Tommaso Pasini, Enrico Santus, Vered Shwartz, Roberto Navigli, and Horacio
  Saggion. 2018.
\newblock {SemEval-2018 Task 9: Hypernym Discovery}.
\newblock In \emph{Proceedings of the 12th International Workshop on Semantic
  Evaluation (SemEval-2018)}, New Orleans, LA, United States. Association for
  Computational Linguistics.

\bibitem[{Cho et~al.(2014)Cho, Merrienboer, Gulcehre, Bahdanau, Bougares,
  Schwenk, and Bengio}]{Cho2014Learning}
Kyunghyun Cho, Bart~Van Merrienboer, Caglar Gulcehre, Dzmitry Bahdanau, Fethi
  Bougares, Holger Schwenk, and Yoshua Bengio. 2014.
\newblock Learning phrase representations using rnn encoder-decoder for
  statistical machine translation.
\newblock In \emph{Proceedings of the 2014 Conference on Empirical Methods in
  Natural Language Processing (EMNLP 2014)}, pages 1724--1734.

\bibitem[{Dagan et~al.(2013)Dagan, Dan, Zanzotto, and
  Sammons}]{Dagan2013Recognizing}
Ido Dagan, Roth Dan, Fabio Zanzotto, and Mark Sammons. 2013.
\newblock Recognizing textual entailment:models and applications.
\newblock \emph{Computational Linguistics}, 41(1):157--160.

\bibitem[{Duchi et~al.(2011)Duchi, Hazan, and Singer}]{Duchi2010Adaptive}
John~C. Duchi, Elad Hazan, and Yoram Singer. 2011.
\newblock Adaptive subgradient methods for online learning and stochastic
  optimization.
\newblock \emph{Journal of Machine Learning Research}, 12(39):2121--2159.

\bibitem[{Espinosa-Anke et~al.(2016{\natexlab{a}})Espinosa-Anke,
  Camacho-Collados, Bovi, and Saggion}]{Espinosa2016Supervised}
Luis Espinosa-Anke, Jose Camacho-Collados, Claudio~Delli Bovi, and Horacio
  Saggion. 2016{\natexlab{a}}.
\newblock Supervised distributional hypernym discovery via domain adaptation.
\newblock In \emph{Proceedings of the 2016 Conference on Empirical Methods in
  Natural Language Processing (EMNLP 2016)}, page 424–435.

\bibitem[{Espinosa-Anke et~al.(2016{\natexlab{b}})Espinosa-Anke, Saggion,
  Ronzano, and Navigli}]{Espinosa2016ExTaSem}
Luis Espinosa-Anke, Horacio Saggion, Francesco Ronzano, and Roberto Navigli.
  2016{\natexlab{b}}.
\newblock Extasem! extending, taxonomizing and semantifying domain
  terminologies.
\newblock In \emph{Proceedings of the Thirtieth AAAI Conference on Artificial
  Intelligence (AAAI-16)}, pages 2594--2600.

\bibitem[{Fountain and Lapata(2012)}]{Fountain2012Taxonomy}
Trevor Fountain and Mirella Lapata. 2012.
\newblock Taxonomy induction using hierarchical random graphs.
\newblock In \emph{Conference of the North American Chapter of the Association
  for Computational Linguistics: Human Language Technologies}, pages 466--476.

\bibitem[{Han et~al.(2013)Han, Kashyap, Finin, Mayfield, and
  Weese}]{Han2013UMBC}
Lushan Han, Abhay Kashyap, Tim Finin, James Mayfield, and Jonathan Weese. 2013.
\newblock Umbc ebiquity-core: Semantic textual similarity systems.
\newblock \emph{Second Joint Conference on Lexical and Computational Semantics
  (*SEM)}, 1:44–52.

\bibitem[{Harabagiu and Hickl(2006)}]{Harabagiu2006Methods}
Sanda Harabagiu and Andrew Hickl. 2006.
\newblock Methods for using textual entailment in open-domain question
  answering.
\newblock In \emph{Proceedings of the 21st International Conference on
  Computational Linguistics and 44th Annual Meeting of the ACL (ACL 2006)},
  pages 905--912.

\bibitem[{Hochreiter and Schmidhuber(1997)}]{Hochreiter1997Long}
Sepp Hochreiter and Jürgen Schmidhuber. 1997.
\newblock Long short-term memory.
\newblock \emph{Neural Computation}, 9(8):1735--1780.

\bibitem[{Kang et~al.(2016)Kang, Haghighi, and Burstein}]{Kang2016TaxoFinder}
Yong~Bin Kang, Pari~Delir Haghighi, and Frada Burstein. 2016.
\newblock Taxofinder: A graph-based approach for taxonomy learning.
\newblock \emph{IEEE Transactions on Knowledge \& Data Engineering},
  28(2):524--536.

\bibitem[{Lei et~al.(2016)Lei, Joshi, Barzilay, Jaakkola, Tymoshenko,
  Moschitti, and Marquez}]{Lei:2016}
Tao Lei, Hrishikesh Joshi, Regina Barzilay, Tommi Jaakkola, Katerina
  Tymoshenko, Alessandro Moschitti, and Lluis Marquez. 2016.
\newblock Semi-supervised question retrieval with gated convolutions.
\newblock In \emph{Proceedings of NAACL-HLT 2016}, pages 1279--1289.

\bibitem[{Li et~al.(2018)Li, Zhang, Ju, and Zhao}]{Li2018}
Haonan Li, Zhisong Zhang, Yuqi Ju, and Hai Zhao. 2018.
\newblock Neural character-level dependency parsing for {Chinese}.
\newblock In \emph{The Thirty-Second AAAI Conference on Artificial Intelligence
  (AAAI-18)}.

\bibitem[{Pelevina et~al.(2016)Pelevina, Arefiev, Biemann, and
  Panchenko}]{pelevina-EtAl:2016:RepL4NLP}
Maria Pelevina, Nikolay Arefiev, Chris Biemann, and Alexander Panchenko. 2016.
\newblock Making sense of word embeddings.
\newblock In \emph{Proceedings of the 1st Workshop on Representation Learning
  for NLP}, pages 174--183.

\bibitem[{Qin et~al.(2016{\natexlab{a}})Qin, Zhang, and Zhao}]{Qin2016Shallow}
Lianhui Qin, Zhisong Zhang, and Hai Zhao. 2016{\natexlab{a}}.
\newblock Shallow discourse parsing using convolutional neural network.
\newblock In \emph{Proceedings of the 54th Annual Meeting of the Association
  for Computational Linguistics (ACL 2016)}, pages 70--77.

\bibitem[{Qin et~al.(2016{\natexlab{b}})Qin, Zhang, and Zhao}]{Qin2016A}
Lianhui Qin, Zhisong Zhang, and Hai Zhao. 2016{\natexlab{b}}.
\newblock A stacking gated neural architecture for implicit discourse relation
  classification.
\newblock In \emph{Conference on Empirical Methods in Natural Language
  Processing (EMNLP 2016)}, pages 2263--2270.

\bibitem[{Qin et~al.(2017)Qin, Zhang, Zhao, Hu, and Xing}]{Qin2017Adversarial}
Lianhui Qin, Zhisong Zhang, Hai Zhao, Zhiting Hu, and Eric~P. Xing. 2017.
\newblock Adversarial connective-exploiting networks for implicit discourse
  relation classification.
\newblock In \emph{Proceedings of the 55th Annual Meeting of the Association
  for Computational Linguistics (ACL 2017)}, pages 1006--1017.

\bibitem[{Roller and Erk(2016)}]{Roller2016Relations}
Stephen Roller and Katrin Erk. 2016.
\newblock Relations such as hypernymy: Identifying and exploiting hearst
  patterns in distributional vectors for lexical entailment.
\newblock In \emph{Proceedings of the 2016 Conference on Empirical Methods in
  Natural Language Processing (EMNLP 2016)}, page 2163–2172.

\bibitem[{Velardi et~al.(2013)Velardi, Faralli, and
  Navigli}]{Velardi2013OntoLearn}
Paola Velardi, Stefano Faralli, and Roberto Navigli. 2013.
\newblock Ontolearn reloaded: A graph-based algorithm for taxonomy induction.
\newblock \emph{Computational Linguistics}, 39(3):665--707.

\bibitem[{Wang et~al.(2017)Wang, Zhao, and Zhang}]{Wang2017A}
Hao Wang, Hai Zhao, and Zhisong Zhang. 2017.
\newblock A transition-based system for universal dependency parsing.
\newblock In \emph{CONLL 2017 Shared Task: Multilingual Parsing From Raw Text
  To Universal Dependencies (CONLL 2017)}, pages 191--197.

\bibitem[{Wang et~al.(2016{\natexlab{a}})Wang, Qian, Soong, He, and
  Zhao}]{Wang2016Learning}
Peilu Wang, Yao Qian, Frank~K. Soong, Lei He, and Hai Zhao. 2016{\natexlab{a}}.
\newblock Learning distributed word representations for bidirectional lstm
  recurrent neural network.
\newblock In \emph{Conference of the North American Chapter of the Association
  for Computational Linguistics: Human Language Technologies (NAACL 2016)},
  pages 527--533.

\bibitem[{Wang et~al.(2016{\natexlab{b}})Wang, Utiyama, Goto, Sumita, Zhao, and
  Lu}]{Wang2016Converting}
Rui Wang, Masao Utiyama, Isao Goto, Eiichiro Sumita, Hai Zhao, and Bao~Liang
  Lu. 2016{\natexlab{b}}.
\newblock Converting continuous-space language models into n-gram language
  models with efficient bilingual pruning for statistical machine translation.
\newblock \emph{ACM Transactions on Asian and Low-Resource Language Information
  Processing}, 15(3):11.

\bibitem[{Wang et~al.(2015)Wang, Zhao, Lu, Utiyama, and
  Sumita}]{Wang2015Bilingual}
Rui Wang, Hai Zhao, Bao~Liang Lu, Masao Utiyama, and Eiichiro Sumita. 2015.
\newblock Bilingual continuous-space language model growing for statistical
  machine translation.
\newblock \emph{IEEE/ACM Transactions on Audio Speech \& Language Processing},
  23(7):1209--1220.

\bibitem[{Wang et~al.(2016{\natexlab{c}})Wang, Zhao, Lu, Utiyama, and
  Sumita}]{Wang2016Connecting}
Rui Wang, Hai Zhao, Bao~Liang Lu, Masao Utiyama, and Eiichro Sumita.
  2016{\natexlab{c}}.
\newblock Connecting phrase based statistical machine translation adaptation.
\newblock In \emph{Proceedings of COLING 2016, the 26th International
  Conference on Computational Linguistics: Technical Papers (COLING 2016)},
  page 3135–3145.

\bibitem[{Yahya et~al.(2013)Yahya, Berberich, Elbassuoni, and
  Weikum}]{Yahya2013Robust}
Mohamed Yahya, Klaus Berberich, Shady Elbassuoni, and Gerhard Weikum. 2013.
\newblock Robust question answering over the web of linked data.
\newblock In \emph{Proceedings of the 22nd ACM international conference on
  Conference on information \& knowledge management (CIKM 2013)}, pages
  1107--1116.

\bibitem[{Zhang et~al.(2016)Zhang, Zhao, and Qin}]{Zhang2016Probabilistic}
Zhisong Zhang, Hai Zhao, and Lianhui Qin. 2016.
\newblock Probabilistic graph-based dependency parsing with convolutional
  neural network.
\newblock In \emph{Proceedings of the 54th Annual Meeting of the Association
  for Computational Linguistics (ACL 2016)}, pages 1382--1392.

\bibitem[{Zhao et~al.(2017{\natexlab{a}})Zhao, Cai, Huang, and Kit}]{Zhao2017A}
Hai Zhao, Deng Cai, Changning Huang, and Chunyu Kit. 2017{\natexlab{a}}.
\newblock \emph{{Chinese} Word Segmentation, a decade review (2007-2017)}.
\newblock {China} Social Sciences Press, Beijing, China.

\bibitem[{Zhao et~al.(2017{\natexlab{b}})Zhao, Cai, Xin, Wang, and
  Jia}]{Zhao2017B}
Hai Zhao, Deng Cai, Yang Xin, Yuzhu Wang, and Zhongye Jia. 2017{\natexlab{b}}.
\newblock A hybrid model for {Chinese} spelling check.
\newblock \emph{ACM Transactions on Asian Low-Resource Language Information
  Process}, pages 1--22.

\end{thebibliography}
\bibliographystyle{acl_natbib}

\appendix
\end{document}